# Image Compression with Iterated Function Systems, Finite Automata and Zerotrees:
# Grand Unification*

Oleg Kiselyov and Paul Fisher

Computer and Information Sciences, Inc.
303 N.Carroll blvd., Suite 108
Denton Texas 76201

*Phone: (817) 484-1165, FAX: (817) 484-0586*
*Email: oleg@ponder.csci.unt.edu, fisher@compsci.com*

Index words: self-similarity, fractal image compression, iterated function system, multiresolutional analysis, wavelet decomposition, predictor/corrector.

## Abstract

Fractal image compression, Culik's image compression and zerotree prediction coding of wavelet image decomposition coefficients succeed only because typical images being compressed possess a significant degree of self-similarity. This is a unifying, common concept of these seemingly dissimilar compression techniques, which may not be apparent due to particular terminologies each of the methods uses. Besides the common concept, these methods turn out to be even more tightly related, to the point of algorithmical reducibility of one technique to another. The goal of the present paper is to demonstrate these relations.

The paper offers a **plain-term** interpretation of Culik's image compression, a very capable yet undeservingly underrepresented method giving spectacular results. The Culik's method will be explained in regular image processing terms, *without* resorting to finite state machines and similar lofty language. The interpretation is shown to be *algorithmically* related to an IFS fractal image compression method: an IFS can be *exactly* transformed into Culik's image code. Using this transformation, we will prove that in a self-similar (part of an) image any zero wavelet coefficient is the root of a zerotree, or its branch.

The paper discusses the zerotree coding of (wavelet/projection) coefficients as a common predictor/corrector, applied *vertically* through different layers of a multiresolutional decomposition, rather than within the same view. This interpretation leads to an insight into the evolution of image compression techniques: from a causal single-layer prediction, to non-causal same-view predictions (wavelet decomposition among others) and to a causal cross-layer prediction (zero-trees, Culik's method). A non-causal cross-level prediction appears to be the next step. Will someone take it?

## I. Introduction

The present paper deals with analysis, generalizations and unifications of the latest group of powerful image compression techniques: fractal image compression with Iterated Function Systems (IFS) [BARN93], Culik's compression with finite automata [CULI95] and Shapiro's embedded coding of wavelet coefficients using zerotrees [SHAP94]. All three techniques achieve premium results by exploiting properties of self-similarity of typical images. In more precise terms, they all rely on the fact that parts of image representations at different resolutions may in some sense be similar. Therefore, a higher-resolution representation may be rather accurately predicted from a low-resolution one. This leads to compression due to compactness of the low-resolution view and smallness of the prediction errors (corrections).

---
* This work was supported in part by US Navy SPAWAR Grant N00039-94-C-0013 "Compression of Geophysical Data" and by US Army Research Office TN 93-461 administered by Battelle Research Office.

Although this conceptual unity is fairly obvious, details of the precise relationship among these methods are a bit obscure. This is partly because of specialized non-intersecting terminology domains used to describe these techniques: iterated transforms, finite automata, wavelet image transform. In the present paper, we will show that all three methods can be formulated in plain terms of a common language, which makes the kinship of these techniques manifest. Furthermore, it turns out that these methods are not only conceptually related, they are algorithmically reducible as well. The paper demonstrates an algorithm by which an image coding with one technique can be *exactly* transformed into another method's image code, with both codes yielding identical reconstruction results. Specifically, we will show how an IFS can be rendered in terms of projection matrices of Culik's method. Although these techniques appear to function in opposite ways (an IFS iteration shrinks the image iterated upon, while a Culik's iteration expands it), the reduction of one to the other is indeed possible, with both iterations producing identical results at all steps up to the final reconstructed image. This transformation also allows us to demonstrate in exact, precise terms how self-similarity of a part of an image gives rise to a zerotree of corresponding wavelet coefficients. In other words, if an image can be adequately represented by an IFS, every zero/insignificant wavelet coefficient in its decomposition is a root of a zerotree branch.

The three methods above can be considered the latest step in evolution of image compression techniques. Since every compressor is based on modelling (prediction) of a source and compact representation (or disregarding) of the prediction errors, what sets different algorithms apart is whether prediction is causal, and what quantity is predicted. For example, CCITT Group III, JPEG lossless, etc., use a causal prediction of a pixel from its same-resolution neighborhood. A Laplacian pyramid decomposition, perfected by a Wavelet image transform, is an example of a non-causal prediction, as first noted by Burt [BURT83]. There, the neighborhood surrounds the pixel in question on *all* flanks. This usually leads to a more accurate prediction (and, therefore, smaller correction). Then came a zerotree coding, a causal prediction of a coefficient/pixel based upon its resolutional neighborhood. This cross-resolutional prediction is indeed causal: if a parent tree node is zero (insignificant), all kid nodes are anticipated to be zeros as well. Non-causal cross-resolution predictor awaits: wavelet decomposition of layers of wavelet decomposition?

## II. Culik's method revealed: fat pixels and exposing projectors

Culik's method is based on an alternative exact representation of an image as a single "fat" pixel, which gets stretched and smeared during repeated expansion operations, until it covers the whole area of the original picture. Unlike a regular, "thin", pixel (which holds a single value: brightness of the corresponding picture element), the fat pixel is a vector. The brightness of the corresponding picture element is computed as a linear combination of the fat pixel vector elements. In the simplest case, one can consider the first element of a fat pixel vector to be a "visible" brightness, with the rest of the vector values being "hidden". The hidden values show up during projection by four matrices, which arrange fat pixel(s) into four quadrants of a larger picture. This representation of an image by a single fat pixel is always possible, and the original image can be reconstructed in its entirety. As an example, the picture below shows a representation of a 4×4 image by a single 16-vector (fat pixel). Different pixels are numbered 1 through 16: these are merely pixel labels rather than actual pixel values.

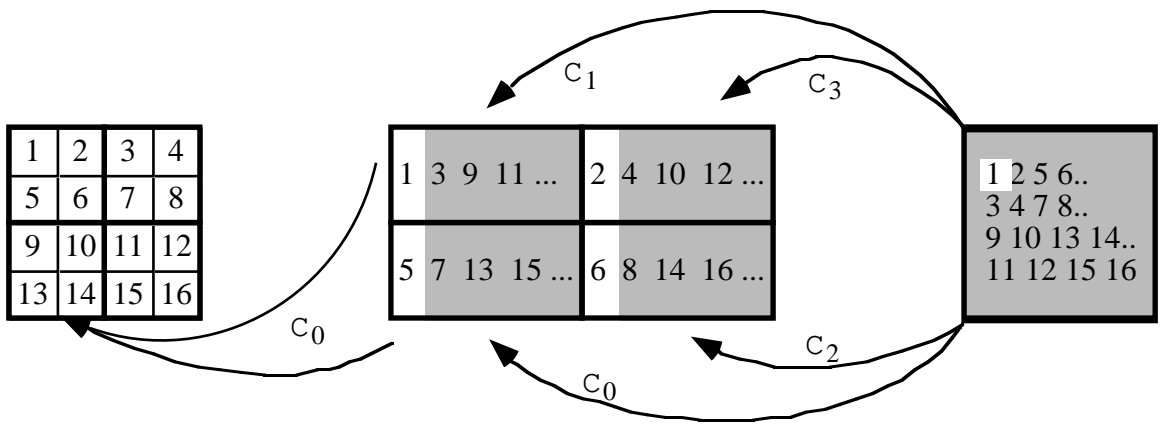

Fig. 1. Example of fat pixel revelations to precisely reconstruct an image. Far left: the original image. Far right: a single fat pixel with 16 components. Center: a partial revelation of hidden components (grayed) upon application of the four transformation matrices.

There are only four projection matrices, $C_0$-$C_3$, which are applied over and over again to produce an image at a finer resolution. For example, applying transform $C_0$ to the original fat pixel (Fig. 1, far right) makes a lower-left fat pixel of a 2×2 square, at the center of Fig. 1. Applying $C_0$ again, to the entire 2×2 square, gives the lower-left quadrant of the image on Fig. 1, far left. The projection matrices in the example above are trivial:

```
1 0 0 0 0 0 0 0 0 0 0 0 0 0 0 0     0 1 0 0 0 0 0 0 0 0 0 0 0 0 0 0     0 0 1 0 0 0 0 0 0 0 0 0 0 0 0 0     0 0 0 1 0 0 0 0 0 0 0 0 0 0 0 0
0 0 0 0 1 0 0 0 0 0 0 0 0 0 0 0     0 0 0 0 0 1 0 0 0 0 0 0 0 0 0 0     0 0 0 0 0 0 1 0 0 0 0 0 0 0 0 0     0 0 0 0 0 0 0 1 0 0 0 0 0 0 0 0
0 0 0 0 0 0 0 0 1 0 0 0 0 0 0 0     0 0 0 0 0 0 0 0 0 1 0 0 0 0 0 0     0 0 0 0 0 0 0 0 0 0 1 0 0 0 0 0     0 0 0 0 0 0 0 0 0 0 0 1 0 0 0 0
0 0 0 0 0 0 0 0 0 0 0 0 1 0 0 0     0 0 0 0 0 0 0 0 0 0 0 0 0 1 0 0     0 0 0 0 0 0 0 0 0 0 0 0 0 0 1 0     0 0 0 0 0 0 0 0 0 0 0 0 0 0 0 1
.........                            .........                            .........                            .........
    $C_1$                                $C_3$                                $C_0$                                $C_2$
```
Fig. 2. Projection matrices for Fig. 1.

They obviously are permutation matrices: matrix **C**$_1$ picks up every forth element of a vector, matrix **C**$_3$ picks the next ones, etc.

The image representation above does not yet give any compression; moreover, we need an additional space to store coefficients of the projection matrices. However, it might turn out that the original image or its close approximation can be reconstructed with less fat pixels. For example, consider a Sierpinski gasket:

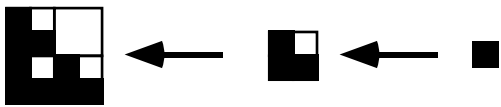

Fig. 3. Making of the Sierpinski gasket: two steps of expanding a thin pixel.

As the figure shows, one needs only a single "thin" pixel and four 1×1 matrices **C**$_0$=**C**$_1$=**C**$_2$=1, **C**$_3$=0 to make the gasket at any resolution. Another example is a diagonal grayscale ramp (this example is almost identical to the one given in Culik's paper [CULI95]). The numbers in squares in the figure below are the pixel values themselves, on a 1-256 scale.

|     |     |
| --- | --- |
| 128 |     |
| 256 |     |
| 128 | 64  |
| 256 | 256 |
| 192 | 128 |
| 256 | 256 |

|     |     |     |     |
| --- | --- | --- | --- |
| 128 | 96  | 64  | 32  |
| 160 | 128 | 96  | 64  |
| 192 | 160 | 128 | 96  |
| 214 | 192 | 160 | 128 |

Fig. 4a. Original fat pixel (a hidden component is grayed)

Fig. 4b. One step of transformation

Fig. 4c. Two steps of transformation (all hidden components have the value of 256, and not shown)

The projection matrices are as follows:

$$C_0 = \begin{pmatrix} \tfrac{1}{2} & \tfrac{1}{2} \\ 0 & 1 \end{pmatrix}, \quad C_1 = C_2 = \begin{pmatrix} \tfrac{1}{2} & \tfrac{1}{4} \\ 0 & 1 \end{pmatrix}, \quad C_3 = \begin{pmatrix} \tfrac{1}{2} & 0 \\ 0 & 1 \end{pmatrix} \qquad (1)$$

One can iterate further to obtain a bigger image, with smoother gray scale gradations. In any case, one needs only a single 2-vector (fat pixel) and four 2×2 matrices, 18 short integers total, to represent even 256×256 and bigger images.

As Culik's presentations at DCC conferences have demonstrated, even realistic pictures (of lenna, among others) can be represented quite compactly with only 300 or so coefficients total (as compared to $^1/_4$M pixels in case of a 512×512 grayscale picture).

## III. Culik's Compression and Iterated Function Systems

Examples of Culik's iterations shown above strongly suggest that Culik's method must be very closely related to iterated function systems. It is indeed: in this section, we will show how one can convert an IFS into Culik's transform/fat pixel.

Iterated Function System (IFS) is a finite collection of contraction mappings [BARN93]. In practice [BARN93, KOMI95], these mappings are usually specified as transformations between two partitionings of the same image into blocks. One, a finer scale partitioning into range blocks, is usually a regular tiling of the image into non-overlapping, usually 4×4 blocks. Another partitioning uses bigger blocks, called domain blocks, which can overlap and do not have to cover the whole picture. Usually domain blocks are twice as big as the range blocks. An IFS is made of separate transformations from a domain block to a range block. A single transformation squeezes the domain block and linearly adjusts its brightness. For example, the figure below depicts a very simple IFS with a single domain block and four smaller range blocks:

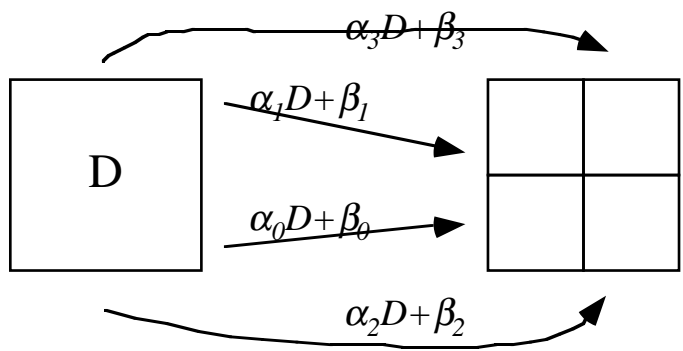

Fig. 5. Sample IFS with a single domain block

Note that the exact sizes of the blocks are irrelevant in this example. The only thing that matters is that the range blocks and the domain block both partition the same image, and that range blocks are half as big in each dimension as compared to a domain block. A linear transform of block's brightness $\alpha D + \beta$ applies to all pixels of the domain block $D$. For example, starting with a square image with a uniform brightness (grayscale) value $y$, and applying the transformations above once, and then again, one obtains:

| | |
|---|---|
| $\alpha_1 y + \beta_1$ | $\alpha_3 y + \beta_3$ |
| $\alpha_0 y + \beta_0$ | $\alpha_2 y + \beta_2$ |

| | | | |
|---|---|---|---|
| $\alpha_1^2 y + \alpha_1\beta_1+\beta_1$ | $\alpha_1\alpha_3 y + \alpha_1\beta_3+\beta_1$ | $\alpha_1\alpha_3 y + \alpha_3\beta_1+\beta_3$ | $\alpha_3^2 y + \alpha_3\beta_3+\beta_3$ |
| $\alpha_0\alpha_1 y + \alpha_1\beta_0+\beta_1$ | $\alpha_1\alpha_2 y + \alpha_1\beta_2+\beta_1$ | $\alpha_0\alpha_3 y + \alpha_3\beta_0+\beta_3$ | $\alpha_2\alpha_3 y + \alpha_3\beta_2+\beta_3$ |
| $\alpha_0\alpha_1 y + \alpha_0\beta_1+\beta_0$ | $\alpha_0\alpha_3 y + \alpha_0\beta_3+\beta_0$ | $\alpha_1\alpha_2 y + \alpha_2\beta_1+\beta_2$ | $\alpha_2\alpha_3 y + \alpha_2\beta_3+\beta_2$ |
| $\alpha_0^2 y + \alpha_0\beta_0+\beta_0$ | $\alpha_0\alpha_2 y + \alpha_0\beta_2+\beta_0$ | $\alpha_0\alpha_2 y + \alpha_2\beta_0+\beta_2$ | $\alpha_2^2 y + \alpha_2\beta_2+\beta_2$ |

Fig. 6a. Application of IFS, Fig. 5, to a square image of uniform brightness *y* considered as a single domain block

Fig. 6b. Application of IFS, Fig. 5, to a an image Fig. 6a considered as a single domain block

The result of one iteration is an image of the same size but with four times as many details. Once the size of a detail diminishes down to one pixel, one may stop iterating: for all practical purposes, "convergence" is achieved. It is obvious that the content of the starting image becomes less and less important, as it shrinks twice at each iteration. Moreover, providing $|\alpha_i| < 1$, all the series on Fig. 6b converge, to a limit not depending on the initial value *y*.

Precisely the same result can be obtained with Culik's transforms, with the original fat pixel **i** and projection matrices as follows:

$$\mathbf{i} = \begin{pmatrix} y \\ 1 \end{pmatrix}, \quad C_k = \begin{pmatrix} \alpha_k & \beta_k \\ 0 & 1 \end{pmatrix}, \quad k = 0, 1, 2, 3 \qquad (2)$$

Indeed, applying the Culik's projection once to the fat pixel **i** gives a picture exactly like Fig. 6a. The only difference is that each quadrant is now a pixel (rather than a square 'subimage'), and it is a fat pixel with a hidden value of 1. Applying the projection once again results in Fig. 6b, with the identical interpretation. In general, it is obvious that an IFS launched from a square image of size $2^m$, and the Culik's transform give identical (and identically sized) results after *m* iterations each.

Note Fig. 4 above is a particular case of this example, Fig. 6 and eq. (2), with

$$\alpha_k = \tfrac{1}{2}, \quad \beta_0 = 128, \beta_1 = \beta_2 = 64, \beta_3 = 0, y = 128 \qquad (3)$$

Let us consider now a more complex IFS, with several domain blocks. First, we will try an example with a single transform, a mapping between a domain and a range block:

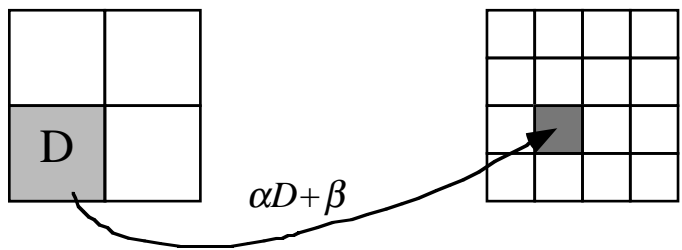

Fig. 7. Sample IFS with a single domain-to-range block mapping

Iterating upon a square with a uniform brightness *y* yields, in turn:

| 0 | 0 | 0 | 0 |
|---|---|---|---|
| 0 | 0 | 0 | 0 |
| 0 | $\alpha y+\beta$ | 0 | 0 |
| 0 | 0 | 0 | 0 |

| 0 | 0 | $\beta$ | $\alpha^2 y + \alpha\beta+\beta$ |
|---|---|---|---|
| 0 | 0 | $\beta$ | $\beta$ |
| 0 | 0 | 0 | 0 |
| 0 | 0 | 0 | 0 |

| 0 | 0 | 0 | 0 | $\beta$ | $\beta$ | $\alpha\beta+\beta$ | $\alpha^3 y+\alpha^2\beta+\alpha\beta+\beta$ |
|---|---|---|---|---|---|---|---|
| 0 | 0 | 0 | 0 | $\beta$ | $\beta$ | $\alpha\beta+\beta$ | $\alpha\beta+\beta$ |
| 0 | 0 | 0 | 0 | $\beta$ | $\beta$ | $\beta$ | $\beta$ |
| 0 | 0 | 0 | 0 | $\beta$ | $\beta$ | $\beta$ | $\beta$ |
| 0 | 0 | 0 | 0 | 0 | 0 | 0 | 0 |
| 0 | 0 | 0 | 0 | 0 | 0 | 0 | 0 |
| 0 | 0 | 0 | 0 | 0 | 0 | 0 | 0 |
| 0 | 0 | 0 | 0 | 0 | 0 | 0 | 0 |

Fig. 8a. Application of IFS, Fig. 7, to a square image of uniform brightness *y* considered covered by the 4 domain blocks

Fig. 8b. Application of IFS, Fig. 7, to the image Fig. 8a considered covered by the 4 domain blocks. Only the lower right quadrant is shown.

Fig. 8c. Application of IFS, Fig. 7, to the image Fig. 8b considered covered by the 4 domain blocks. Only the lower right quadrant is shown.

The corresponding Culik's transformation is:

$$\mathbf{i} = \begin{pmatrix} 0 \\ y \\ 1 \\ 1 \end{pmatrix}, \quad C_0 = \begin{pmatrix} 0 & 1 & 0 & 0 \\ 0 & 0 & 0 & 0 \\ 0 & 0 & 1 & 0 \\ 0 & 0 & 0 & 1 \end{pmatrix}, C_1 = C_2 = \begin{pmatrix} 0 & 0 & 0 & 0 \\ 0 & 0 & 0 & 0 \\ 0 & 0 & 1 & 0 \\ 0 & 0 & 0 & 1 \end{pmatrix}, C_3 = \begin{pmatrix} 0 & 0 & 0 & 0 \\ 0 & \alpha & 0 & \beta \\ 0 & 0 & 1 & 0 \\ 0 & 0 & 0 & 1 \end{pmatrix} \quad (4)$$

It is evident that the first iteration of projecting the starting fat pixel **i** leads to a picture on the right-hand side of Fig. 7; the second iteration results in Fig. 8a. Iterating once more gives, in turn, Fig. 8b and Fig. 8c, etc.

A more complex example involves two domain blocks and two "mutually dependent" transforms:

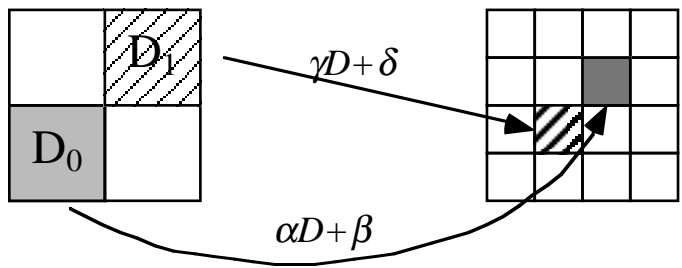

Fig. 9. Sample IFS with a mutually-dependent domain-to-range block mapping

which converges to something like

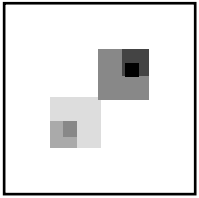

Fig. 10. Third iteration of IFS Fig. 7

The fat pixel **i** and the projection matrices corresponding to the example are as follows:

$$\mathbf{i} = \begin{pmatrix} 0 \\ x \\ y \\ 1 \\ 1 \\ 1 \end{pmatrix}, \quad C_0 = \left( \begin{array}{cccccc} 0 & 1 & 0 & 0 & 0 & 0 \\ 0 & 0 & 0 & 0 & 0 & 0 \\ 0 & \alpha & 0 & 0 & \beta & 0 \\ \hline \mathbf{0} & & & \mathbf{I} & & \end{array} \right), \quad C_1 = C_2 = \left( \begin{array}{cccccc} 0 & 0 & 0 & 0 & 0 & 0 \\ 0 & 0 & 0 & 0 & 0 & 0 \\ 0 & 0 & 0 & 0 & 0 & 0 \\ \hline \mathbf{0} & & & \mathbf{I} & & \end{array} \right), \quad C_3 = \left( \begin{array}{cccccc} 0 & 0 & 1 & 0 & 0 & 0 \\ 0 & 0 & \gamma & 0 & 0 & \delta \\ 0 & 0 & 0 & 0 & 0 & 0 \\ \hline \mathbf{0} & & & \mathbf{I} & & \end{array} \right) \quad (5)$$

where **I** is a 3×3 unit matrix and **0** is a 3×3 zero matrix. As one can easily verify by applying these projection matrices to the fat pixel **i** over and over again, the result at each iteration is identical to that for the IFS Fig. 9.

Thus the IFS and Culik's image compression methods are indeed very tightly related, despite outward differences: an IFS iteration shrinks and reshuffles input image tiles, while a Culik's iteration merely rearranges its input, always in the same regular way. The Culik's method starts with a single domain block (that is, the entire image, or a "fat" pixel), and uses "range" blocks of the same size as the domain block itself. However, the Culik's method makes up for the lost "translational" degrees of freedom by using "fat" pixels and a more complex luminance transform: although linear, but vector rather than scalar. Similar to IFS, Culik's matrices are required not to amplify pixels' luminance [CULI95]; i.e., the matrices should be contractive, or at least, not expanding. Finally, as examples above show, an IFS can indeed be algorithmically reduced to a Culik's transform. The general algorithm of this reduction and its inverse are discussed in more detail in the paper.

It is obvious from the examples above that an IFS with *k* transforms requires a *2(k+1)*-element fat pixel and four *2(k+1)×2(k+1)* projection matrices. Note that the bottom half of these matrices is just a unit matrix, which does not have to be stored at all. The upper halves are also very sparse, which ought to be taken advantage of. For example, one can regard a projection matrix as a distance matrix for a directed weighted graph. Since the matrix is sparse, the corresponding graph would have rather few edges, which can be more efficiently stored as a list. Thus we come to exactly the same weighted graphs Culik originally used to represent his finite automata [CULI95]. Hence, the automata described by Culik are nothing but a neat trick to efficiently store and use sparse projection matrices.

## IV. An IFS image has a zerotree of wavelet coefficients

The title of the section is actually a formulation of a theorem the paper presents and proves. In precise terms, within an image or a part of it with a property of self-similarity, i.e., which can be adequately described/reproduced by an IFS, a zero wavelet coefficient is always a root of a zerotree branch. In other words, if a wavelet coefficient at some particular resolution turns out to be zero (exactly or within some tolerance), all the child coefficients, at finer resolutions, will be zero as well (exactly or within the same tolerance). Thus, as long as an image has enough self-similarity to allow efficient compression by an IFS (or, which is the same, by Culik's method), a zerotree coding of wavelet coefficients would be beneficial. This is the unifying idea mentioned above; the theorem gives it a more precise meaning.

Because of space constraints, we will show the proof of the theorem on a small but characteristic example. We will analyze a case of a simple Haar wavelet transform, which has very short wavelet filters, spanning, in 2D, over a cluster of four "pixels". Consider a set of zoomed-out views of a self-similar (part of a) picture, and assume that the top view is made of the 4-pixel cluster. Following the premise of self-similarity, all these views are well described by an IFS, or (which is the same as we saw above) by a Culik's transform. Let corresponding "fat" pixels of the cluster be $F_0$, $F_1$, $F_2$, and $F_3$, and the projection matrices $C_0$-$C_3$. Let us arrange the four-pixel neighborhood in a block-vector $\mathbf{F}=(F_0\ F_1\ F_2\ F_3)'$. Note that the fat pixels are vectors themselves, that is why we call $\mathbf{F}$ a block-vector. Finer resolution views of the cluster can be obtained by projecting it with (block) matrices $\mathbf{C_i}$. For example, the lower-left quadrant of the cluster at a higher resolution can be computed as

$$\begin{pmatrix} C_0 & 0 & 0 & 0 \\ 0 & C_0 & 0 & 0 \\ 0 & 0 & C_0 & 0 \\ 0 & 0 & 0 & C_0 \end{pmatrix} \begin{pmatrix} F_0 \\ F_1 \\ F_2 \\ F_3 \end{pmatrix} \equiv \mathbf{C_0 F} \qquad (6)$$

where $\mathbf{C_0}$ is a block-projection matrix. The pixels $F_0$-$F_3$ can be combined to yield a (fat) wavelet coefficient $\mathbf{W}$, by using a 2D (high-pass) filter with coefficients $[h_0, h_1, h_2, h_3]$:

$$\mathbf{W} = \mathbf{HF} \equiv \begin{pmatrix} h_0 I & h_1 I & h_2 I & h_3 I \\ h_0 I & h_1 I & h_2 I & h_3 I \\ h_0 I & h_1 I & h_2 I & h_3 I \\ h_0 I & h_1 I & h_2 I & h_3 I \end{pmatrix} \mathbf{F} \equiv \begin{pmatrix} I \\ I \\ I \\ I \end{pmatrix} (h_0 I \ \ h_1 I \ \ h_2 I \ \ h_3 I) \mathbf{F} \qquad (7)$$

where $I$ is a unit matrix of the size that of $F_i$. The pivoting point of the proof is the fact that matrix $\mathbf{H}$ is commutative with $\mathbf{C_i}$. This is easy to see by directly computing $\mathbf{C_i H}$ and $\mathbf{HC_i}$, which in both cases gives:

$$\mathbf{C}_i \mathbf{H} \equiv \mathbf{H} \mathbf{C}_i \equiv \begin{pmatrix} 1 \\ 1 \\ 1 \\ 1 \end{pmatrix} \begin{pmatrix} h_0 C_i & h_1 C_i & h_2 C_i & h_3 C_i \end{pmatrix} \qquad (8)$$

A child wavelet coefficient at any finer resolution can be computed then as

$$\mathbf{W}_{kid} = \mathbf{H} \mathbf{C}_{i_0} \mathbf{C}_{i_1} \mathrm{L} \ \mathbf{C}_{i_k} \mathbf{F} = \mathbf{C}_{i_0} \mathbf{C}_{i_1} \mathrm{L} \ \mathbf{C}_{i_k} \mathbf{H} \mathbf{F} \qquad (9)$$

Note that the first element of a fat pixel $F_i$, $F_{i_0}$, is the visible pixel. The hidden elements are either 1, or can be set to 1, because it does not matter in the case of contracting matrices $\mathbf{C}_i$, as we saw above. Since the wavelet filter $\mathbf{H}$ is high-pass, $(h_0\ h_1\ h_2\ h_3)(1\ 1\ 1\ 1)'$ is exactly zero. Therefore, if the wavelet-filtering of visible pixels $F_{i_0}$ gives zero as well, the entire fat wavelet coefficient $\mathbf{W}=\mathbf{HF}$ is a zero matrix. It follows then from eq. (9) that all the children wavelet coefficients are zeros as well.

One can easily accommodate other wavelet filters by considering larger neighborhood of pixels. Block-vector $\mathbf{F}$ and block-matrices $\mathbf{C}_i$ would have more block-rows/columns, but the derivations remain the same. It is also easy to generalize the result to a case when a wavelet coefficient is not exactly zero, but small. As long as matrices $\mathbf{C}_i$ are not expanding (that is, convergence is guaranteed), all kid wavelet coefficients would be just as small as their parent.

**References**

[BARN93]  Barnsley, M.F., Hurd L.P., Fractal Image Compression, AK Peters Ltd., 1993, 244 p.

[BURT83]  Burt, P., Adelson, E., "The Laplacian Pyramid as a Compact Image Code," *IEEE Trans. Comm.*, Vol. 31, No. 4, pp. 532-540, April 1983.

[CULI95] Culik, Karel II, Kari, J., "Finite State Methods for Compression and Manipulation of Images," in *Proc. DCC'95, 1995 Data Compression Conference*, Snowbird, Utah, pp.142-151, March 28-30, 1995.

[KOMI95] Kominek, John, "Convergence of Fractal Encoded Images," in *Proc. DCC'95, 1995 Data Compression Conference*, Snowbird, Utah, pp.242-251, March 28-30, 1995.

[SHAP94] Shapiro, J.M., "Embedded image coding using zerotrees of wavelet coefficients," *IEEE Trans. on Signal Processing*, Vol. 41, No. 12, pp. 3445-3462, December 1994.